\documentclass[conference]{IEEEtran}
\IEEEoverridecommandlockouts

\usepackage[dvipsnames,table,xcdraw]{xcolor}
\usepackage[pagebackref=true,breaklinks=true,colorlinks,citecolor=ForestGreen]{hyperref}
\usepackage{ragged2e}

\usepackage{cite}
\usepackage{amsmath,amssymb,amsfonts}
\usepackage{algorithmic}
\usepackage{graphicx}
\usepackage{textcomp}

\usepackage{algorithm}
\usepackage{epsfig}
\usepackage{booktabs}
\usepackage{colortbl}
\usepackage{multirow}
\usepackage{pifont}
\usepackage[switch]{lineno}
\usepackage{hyperref}
\usepackage{ragged2e}
\usepackage{tabularx}
\usepackage{enumitem}

\usepackage{pgfplots}
\usepackage{tikz}
\pgfplotsset{compat=1.18}

\def\BibTeX{{\rm B\kern-.05em{\sc i\kern-.025em b}\kern-.08em
    T\kern-.1667em\lower.7ex\hbox{E}\kern-.125emX}}
\begin{document}

\title{DisentTalk: Cross-lingual Talking Face Generation via Semantic Disentangled Diffusion Model\thanks{$^{\star}$ Corresponding authors.}
\thanks{This work was supported by NSFC under 62272456.}
\thanks{We extend our special thanks to Ruoyu Chen for helping revise Figures 1, the Introduction, and the Abstract.}
}
\author{
\IEEEauthorblockN{Kangwei Liu$^{1,2}$, Junwu Liu$^{1,2}$, Yun Cao$^{1,2}$, Jinlin Guo$^{3}$, Xiaowei Yi$^{1,2,\star}$}
\IEEEauthorblockA{$^1$ Institute of Information Engineering, Chinese Academy of Sciences, Beijing 100085, China}
\IEEEauthorblockA{$^2$ School of Cyber Security, University of Chinese Academy of Sciences, Beijing 100085, China} 
\IEEEauthorblockA{$^3$ Laboratory for Big Data and Decision, School of System Engineering, National University of Defense} 
\IEEEauthorblockA{\{liukangwei, liujunwu, caoyun, yixiaowei\}@iie.ac.cn, gjlin99@nudt.edu.cn}
}

\maketitle

\begin{abstract}
Recent advances in talking face generation have significantly improved facial animation synthesis. However, existing approaches face fundamental limitations: 3DMM-based methods maintain temporal consistency but lack fine-grained regional control, while Stable Diffusion-based methods enable spatial manipulation but suffer from temporal inconsistencies. The integration of these approaches is hindered by incompatible control mechanisms and semantic entanglement of facial representations. This paper presents DisentTalk, introducing a data-driven semantic disentanglement framework that decomposes 3DMM expression parameters into meaningful subspaces for fine-grained facial control. Building upon this disentangled representation, we develop a hierarchical latent diffusion architecture that operates in 3DMM parameter space, integrating region-aware attention mechanisms to ensure both spatial precision and temporal coherence. To address the scarcity of high-quality Chinese training data, we introduce CHDTF, a Chinese high-definition talking face dataset. Extensive experiments show superior performance over existing methods across multiple metrics, including lip synchronization, expression quality, and temporal consistency. Project Page: \url{https://kangweiiliu.github.io/DisentTalk}.
\end{abstract}


\section{Introduction}
Recent advancements in talking face generation models \cite{makeittalk, avct, audio2head, sadtalker, styletalk, iplap,wav2lip, pcavs, pdfgc, latentdiffu, difftalk, emo, aniportrait, hallo, dreamtalk} have significantly enhanced the quality of facial animation synthesis. However, achieving natural and temporally coherent talking face videos remains challenging, particularly in cross-lingual scenarios \cite{styletalk}.

\begin{figure}[!t]
  \centering
  \includegraphics[width=\columnwidth]{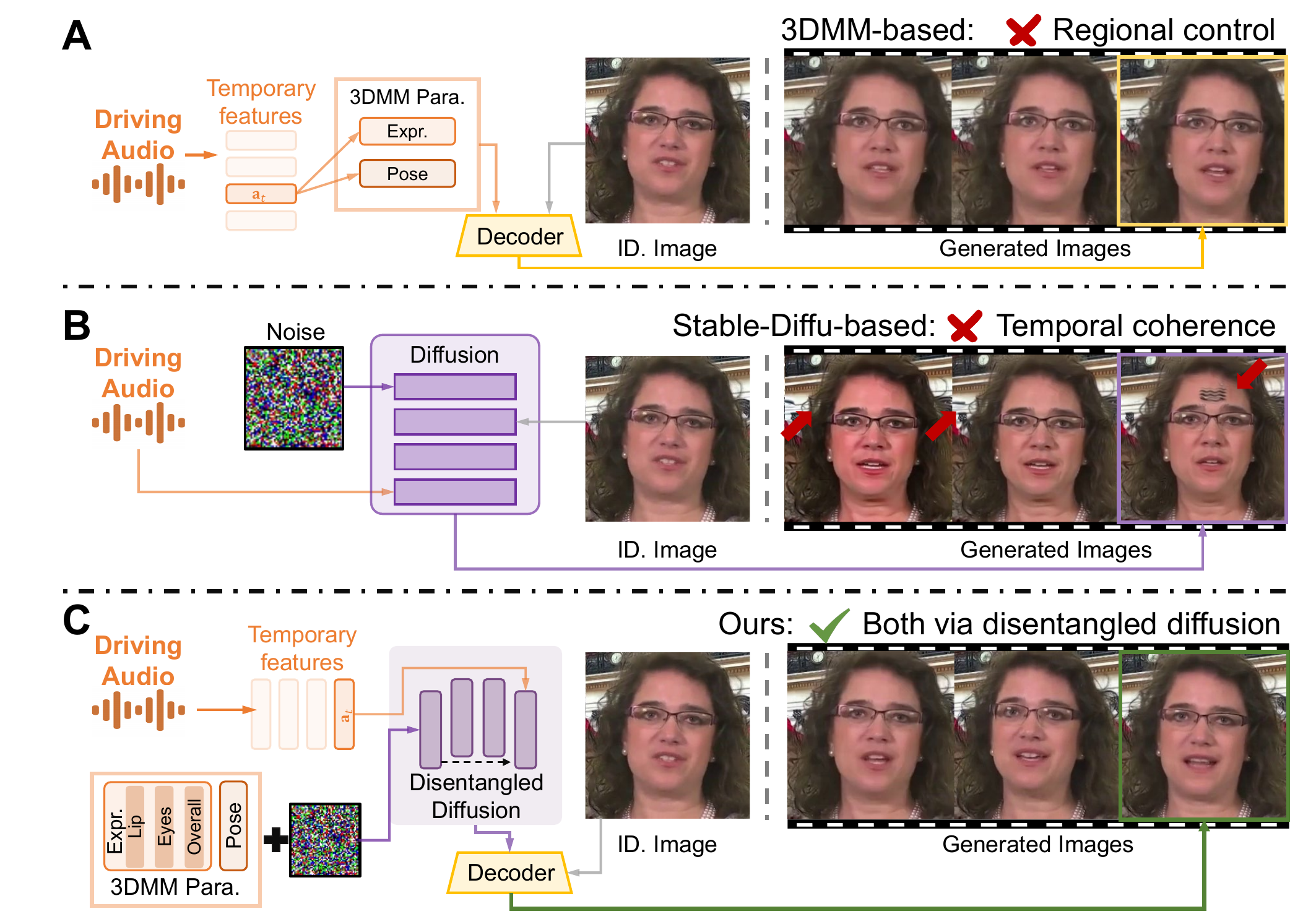}
  \vspace{-3mm}
  \caption{Comparison of talking face generation approaches. (A) 3DMM-based methods: temporally consistent but lacks regional control. (B) Stable diffusion-based methods enable spatial control but are temporally inconsistent. (C) Our DisentTalk: achieves both through the disentangled latent diffusion model.}
  \label{fig:abstrac_fig}
  \vspace{-3mm}
\end{figure}

Current approaches primarily fall into two categories, each with fundamental limitations. As shown in Fig. \ref{fig:abstrac_fig}, 3DMM-based methods \cite{sadtalker, styletalk, dreamtalk} employ parametric models \cite{3dmm, 3drecon} to maintain temporal consistency, but suffer from semantic entanglement in their expression parameters, limiting fine-grained control over distinct facial regions. While Stable Diffusion-based methods \cite{aniportrait, hallo} enable direct spatial manipulation through geometric controls, their frame-by-frame generation nature leads to temporal inconsistencies. The integration of these approaches remains challenging due to their incompatible control mechanisms: Stable Diffusion \cite{latentdiffu} operates on geometric signals (e.g., landmarks), while 3DMM \cite{3dmm} represents facial expressions through sparse coefficients.

To address these challenges, we present DisentTalk, a latent diffusion framework operating in 3DMM parameter space that enables both temporal consistency and fine-grained control. To address the semantic entanglement of 3DMM parameters, we propose a data-driven disentanglement strategy that decomposes 3DMM expression parameters into semantic subspaces, bridging the gap between geometric control and parametric representation. This decomposition enables control over lip articulation, eye dynamics, and global expressions while preserving the properties of 3DMM representation. The framework consists of a hierarchical latent diffusion architecture that integrates spatial and temporal features during the generation process. Through region-aware spatial attention and multiscale temporal modeling, the architecture is able to capture facial details and motion patterns. To enrich the training data and enable cross-lingual evaluation, we introduce CHDTF, a Chinese high-definition talking face dataset. The main contributions are summarized as follows:

\begin{itemize}
    \item A semantic disentanglement framework for 3DMM parameters that enables control over distinct facial regions while preserving temporal relationships.
    \item A hierarchical latent diffusion architecture in 3DMM parameter space with region-aware attention mechanisms for spatial-temporal feature modeling.
    \item A Chinese high-definition talking face dataset (CHDTF) substantially enriches the training data for talking face generation.
    \item Extensive experiments showing superior performance over existing methods across multiple metrics, including lip synchronization, expression quality, and temporal consistency.
\end{itemize}

\begin{figure*}[!ht]
  \centering
  \includegraphics[width=0.95\textwidth]{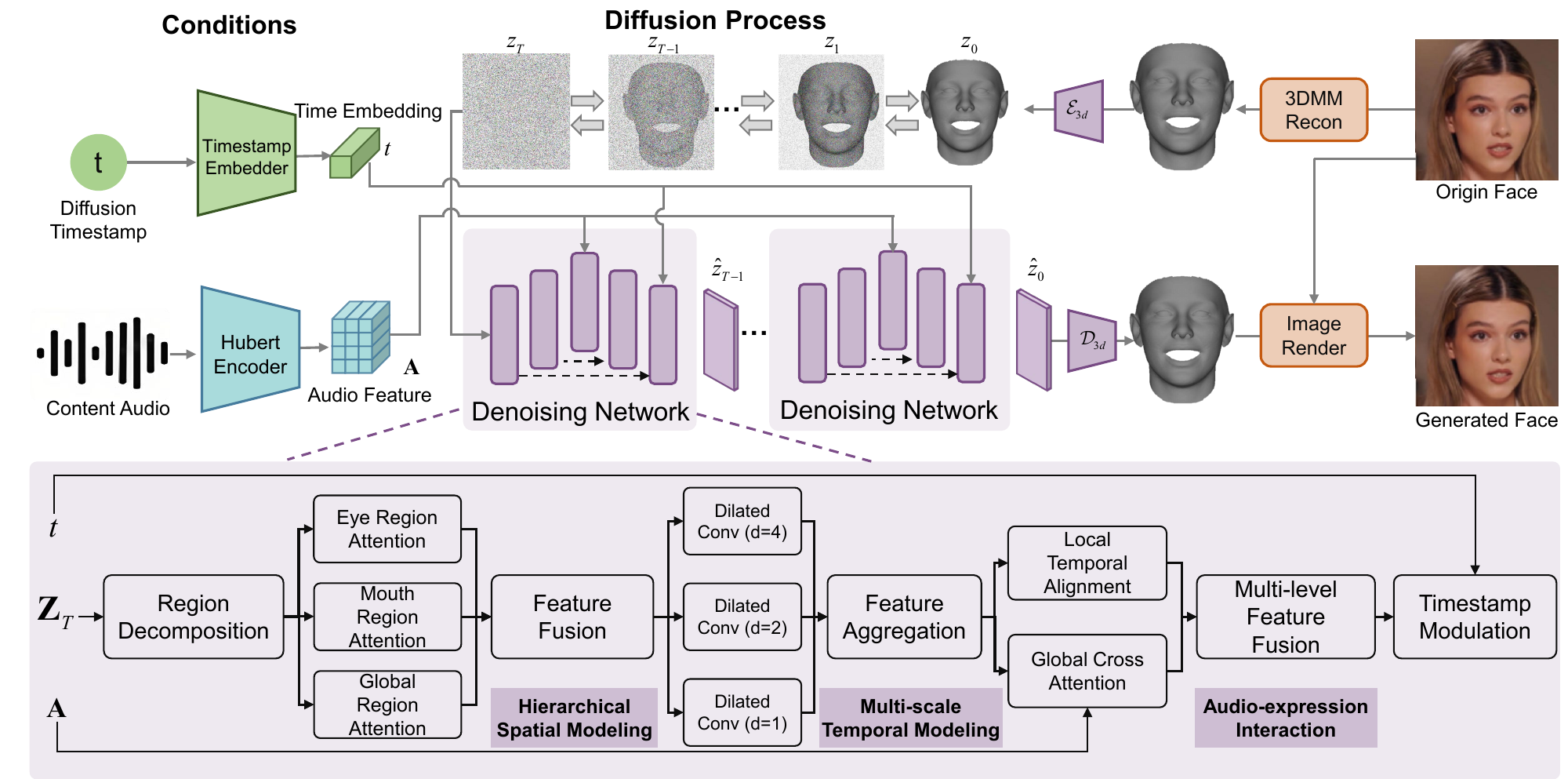}
  \caption{Overview of our spatio-temporal aware diffusion framework. The model leverages disentangled 3DMM parameters and hierarchical attention mechanisms to achieve precise control over distinct facial regions while maintaining temporal coherence.}
  \label{fig:overview}
\end{figure*}
\vspace{-2mm}
\section{Related Work}

\noindent\textbf{Deterministic Methods.}
Early approaches can be categorized into end-to-end GAN-based methods (e.g., Wav2Lip~\cite{wav2lip}, PCAVS~\cite{pcavs}, \cite{produb}) and intermediate representation-based methods using 3DMM parameters \cite{sadtalker, styletalk} or landmarks \cite{makeittalk, iplap}. While these methods achieve reasonable lip synchronization and temporal consistency, they struggle to capture the probabilistic nature of speech-driven facial expressions, resulting in limited expressiveness.

\noindent\textbf{Diffusion-based Methods.}
Recent diffusion-based methods have shown promising results in enhancing generation quality. Stable Diffusion-based approaches (EMO~\cite{emo}, DiffTalk~\cite{difftalk}) excel at spatial detail modeling but suffer from temporal inconsistency and high computational costs. DreamTalk~\cite{dreamtalk} combines 3DMM with diffusion models to improve temporal coherence, yet its holistic parameter modeling limits fine-grained facial control.

\noindent\textbf{Cross-lingual Generation.}
Current datasets and methods primarily focus on English content \cite{hdtf, vox2}, with limited success in cross-lingual scenarios. This highlights the need for Chinese high-definition talking face datasets to advance cross-lingual talking face generation.

\section{Method}
\vspace{-1mm}
We propose a novel framework for audio-driven facial animation with two key components: (1) semantic disentanglement of 3DMM expression parameters, and (2) a spatio-temporal aware diffusion model leveraging this disentangled representation. Fig. \ref{fig:overview} illustrates our pipeline.

\vspace{-1mm}
\subsection{3DMM Expression Parameter Disentanglement}
\begin{figure}[!t]
  \centering
  \includegraphics[width=\columnwidth]{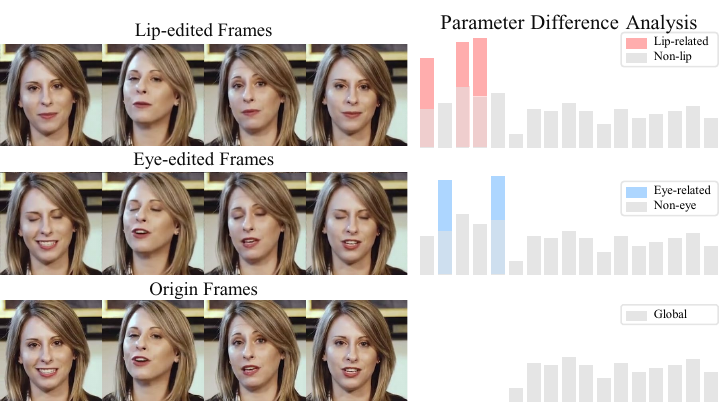}
  \caption{3DMM parameter disentanglement process. We decompose expression parameters into lip articulation, eye dynamics, and global expression subspaces through data-driven analysis of facial modifications.}
  \label{fig:3dmm_disentangle}
\end{figure}

While 3DMM effectively encodes facial expressions through PCA of 3D scans \cite{3dmm,3drecon}, its mathematically orthogonal dimensions often entangle semantic facial movements, limiting precise control in audio-driven animation. To address this, we propose a data-driven disentanglement framework that decomposes 3DMM parameters into three semantically meaningful subspaces: lip articulation, eye dynamics, and global expressions (Fig. \ref{fig:3dmm_disentangle}).

To identify region-specific dimensions, we transform source videos into states with closed lips and eyes using a facial editing model \cite{liveportrait}. By analyzing the parameter differences between original and modified states, we select dimensions that are most sensitive to each regional movement:
\begin{equation}
    D_r = \arg \max_{K_r} \{\frac{|\Delta_r^i|}{\sigma_i}\}_{i=1}^N,~~~~r \in \{lip, eye\},
\end{equation}
where $\Delta_r^i$ denotes the parameter difference in dimension $i$ for region $r$, $\sigma_i$ normalizes the variation scale, and $K_r$ represents the number of selected dimensions. The remaining dimensions form the global expression subspace:
\begin{equation}
    D_{global} = \{1,...,N\} \setminus (D_{lip} \cup D_{eye}).
\end{equation}

This decomposition preserves 3DMM's mathematical properties while introducing semantic interpretability, enabling fine-grained control over facial regions. The disentangled representation serves as the foundation for our attention-based architecture, facilitating more natural facial animations.



\vspace{-1mm}
\subsection{Spatio-temporal Aware Diffusion Model}
\vspace{-1mm}
Building upon our disentangled 3DMM representation, we propose a spatio-temporal aware denoising architecture for audio-driven facial animation. Unlike previous methods that process spatial and temporal features in isolation \cite{dreamtalk, hallo}, our approach maintains joint awareness of both domains throughout the generation process. The model takes the 3DMM expression parameters $\mathbf{E}$, diffusion timestep $t$, and hubert-based \cite{hubert} audio features $\mathbf{A}$ as input.

\noindent \textbf{Latent Diffusion Model.} We adopt a latent diffusion model \cite{latentdiffu} operating in the 3DMM motion space. Given a latent facial sequence $\mathbf{Z}_0$ from $\mathbf{E}$, the forward process gradually adds Gaussian noise over $T$ timesteps:
\begin{equation}
q(\mathbf{Z}_t \mid \mathbf{Z}_{0}) = \mathcal{N}(\sqrt{1 - \beta_t} \mathbf{Z}_{0}, \beta_t \mathbf{I}),
\end{equation}
where $\beta_t$ controls the noise schedule. The denoising process leverages our spatio-temporal aware network to progressively recover the original motion $\mathbf{Z}_0$ from a Gaussian noise $\mathbf{Z}_N \sim \mathcal{N}(0, \mathbf{I})$.

\noindent \textbf{Denoising Network Architecture.} Our denoising network comprises three key components that jointly process spatial and temporal information: (1) hierarchical region-aware spatial attention for region-specific feature extraction, (2) multi-scale temporal convolution for temporal dynamics modeling, and (3) enhanced audio-expression cross-attention for audio-visual synchronization.

\noindent \textbf{Hierarchical Region-aware Spatial Attention.} 
Leveraging our semantic decomposition of 3DMM parameters, we first introduce a region-aware spatial attention mechanism that maintains distinct facial dynamics while preserving their temporal relationships. This mechanism extends beyond conventional spatial attention by incorporating temporal context into region-specific processing:
\begin{equation}
    \mathbf{F}_{spatial} = \mathbf{W}_f[\text{Attention}(\mathbf{E}_r)]_{r \in \mathcal{R}},
\end{equation}
where $\mathbf{E}_r$ represents region-specific parameters,  $\mathbf{W}_f$ is the feature fusion matrix, $\mathcal{R}=\{eye, lip, global\}$ denotes the set of facial regions, and $[\cdot]_{r \in \mathcal{R}}$ represents concatenation over all regions. This formulation enables the model to capture spatially localized features.  

\noindent \textbf{Multi-scale Temporal Convolution.} 
To complement the spatial attention while preserving fine-grained temporal dynamics, we develop a hierarchical temporal processing framework that maintains spatial awareness. Our design employs spatially-aware depth-wise separable convolutions with exponentially increasing temporal receptive fields:
\begin{equation}
    \mathbf{F}_{temp} = \mathcal{F}_{pw}(\text{Concat}[\{\mathcal{F}_{dw}^i(\mathbf{F}_{spatial})\}_{i=0}^2]),
\end{equation}
where $\mathcal{F}_{dw}^i$ denotes depth-wise convolution with dilation rate $2^i$, and $\mathcal{F}_{pw}$ represents point-wise convolution for feature aggregation. This multi-scale architecture preserves spatial relationships while modeling temporal dependencies at different scales.

\noindent \textbf{Enhanced Audio-Expression Cross-attention.} 
To achieve unified spatio-temporal synchronization with audio features, we propose a dual-path cross-attention mechanism that simultaneously processes local temporal correlations and global audio-visual context:
\begin{equation}
    \mathbf{F}_{final} = \mathbf{W}_o[\mathcal{F}_{conv}(\mathbf{F}_{temp}); \mathcal{F}_{attn}(\mathbf{F}_{temp}, \mathbf{A})] + b_o,
\end{equation}
where $\mathcal{F}_{conv}$ is a local convolution with window size $k$ maintaining temporal correlations, $\mathcal{F}_{attn}$ captures global audio-visual relationships through causal cross-attention, and $\mathbf{W}_o$ is the multi-level feature fusion matrix. The framework is guided by time-dependent FiLM \cite{film} conditioning:
\begin{equation}
    \mathbf{F}_{out} = \gamma(t)\mathbf{F}_{final} + \beta(t),
\end{equation}
where $\gamma(t)$ and $\beta(t)$ modulate features throughout the diffusion process, ensuring coherent spatio-temporal generation.

\noindent \textbf{Training Strategy.} Our training framework combines classifier-free guidance \cite{classfree} with a novel 3DMM-based lip synchronization constraint. During training, audio features are randomly masked (p=0.1) for conditional and unconditional generation, with noise estimation loss:
\begin{equation} 
\mathcal{L}_{\text{noise}} = \mathbb{E}_{\mathbf{Z}_0, t, \epsilon} \left[ \left\| \epsilon - \epsilon_{\theta}(\mathbf{Z}_t, t, \mathbf{A}) \right\|^2 \right],
\end{equation} 
and inference guided by:
\begin{equation} 
\epsilon^{*}_{n} = s \epsilon_{\theta}(\mathbf{Z}_n, n, \mathbf{A}) + (1 - s) \epsilon_{\theta}(\mathbf{Z}_n, n, \varnothing).
\end{equation}

\noindent \textbf{3DMM-based Lip Sync Loss.} Unlike previous SyncNet-based \cite{wav2lip} approaches using image-level features, we propose a parameter-level synchronization loss operating directly on disentangled lip parameters. Given HuBERT features $\mathbf{A}$ and lip motion parameters $\mathbf{E}_m$, we employ two parallel encoders (details in our appendix) to extract normalized embeddings for synchronization assessment:
\begin{equation}
\mathcal{L}_{sync} = \text{BCELoss}(\text{cos}(\frac{f_a(\mathbf{A})}{\|f_a(\mathbf{A})\|_2}, \frac{f_m(\mathbf{E}_m)}{\|f_m(\mathbf{E}_m)\|_2}), y),
\end{equation}
where $f_a$ and $f_m$ are audio and mouth parameter encoders respectively, and $y \in \{0,1\}$ indicates synchronization. The final objective is $\mathcal{L}_{\text{total}} = \mathcal{L}_{\text{noise}} + \lambda \mathcal{L}_{\text{sync}}$. After getting the denoised 3DMM parameters, we use a 3DMM-based face render \cite{pir} to generate the final talking face video.

\begin{table*}[t!]
  \centering
  \caption{Quantitative comparison on \texttt{HDTF} and \texttt{CHDTF} datasets. Methods are categorized into deterministic and diffusion-based approaches. Eye Blink indicates motion source (Generated/Handcraft/From ref). \textbf{Bold} and \underline{underlined} denote Top-2 results.}
  \resizebox{\textwidth}{!}{
    \begin{tabular}{llcccccc|cccccc}
      \toprule
      \multirow{2}{*}{Category} & \multirow{2}{*}{Methods} & \multirow{2}{*}{Eye Blink} & \multicolumn{5}{c}{\texttt{HDTF}\cite{hdtf}} & \multicolumn{5}{c}{\texttt{CHDTF}} \\
      \cmidrule(lr){4-8} \cmidrule(lr){9-13}
      & & & LSE-C$\uparrow$ & LMD$\downarrow$ & Blink$\uparrow$ & CPBD$\uparrow$ & PSNR$\uparrow$ & LSE-C$\uparrow$ & LMD$\downarrow$ & Blink$\uparrow$ & CPBD$\uparrow$ & PSNR$\uparrow$ \\
      \midrule
      Reference & GT & - & 8.316 & 0.000 & 0.198 & 0.906 & inf & 7.183 & 0.000 & 0.223 & 0.851 & inf \\
      \midrule
      \multirow{7}{*}{{Deterministic-based}} 
      & AVCT \cite{avct} & Generated & 5.052 & 3.262 & 0.082 & 0.688 & 17.182 & 1.813 & 3.246 & 0.088 & 0.831 & 14.540 \\
      & Audio2Head \cite{audio2head} & Generated & 6.779 & 2.142 & 0.098 & 0.729 & 17.714 & 5.502 & 2.297 & 0.023 & 0.796 & 14.633 \\
      & MakeItTalk \cite{makeittalk} & Hand craft & 4.776 & 2.704 & 0.201 & 0.655 & 19.027 & 4.548 & 2.809 & 0.202 & 0.780 & 15.625 \\
      & SadTalker \cite{sadtalker} & Hand craft & 6.558 & 2.556 & 0.082 & 0.696 & 16.518 & 6.112 & 2.315 & 0.078 & 0.798 & 14.512 \\
      & Styletalk \cite{styletalk} & Generated & 2.073 & 1.732 & 0.082 & 0.892 & 25.067 & 0.561 & 3.612 & 0.002 & 0.839 & 13.913 \\
      & IPLAP \cite{iplap} & From ref & 5.796 & 1.570 & 0.178 & 0.831 & 33.544 & 4.592 & 1.944 & 0.172 & 0.851 & 32.094 \\
      & PDFGC \cite{pdfgc} & From ref & 6.682 & 3.508 & 0.192 & 0.547 & 13.752 & 5.891 & 3.210 & 0.287 & 0.715 & 10.493 \\
      \midrule
      \multirow{5}{*}{Diffusion-based} 
      & Dreamtalk \cite{dreamtalk} & Generated & \underline{7.140} & \textbf{1.831} & 0.038 & 0.709 & 20.062 & 6.017 & 2.919 & 0.000 & 0.724 & 13.786 \\
      & Aniportrait \cite{aniportrait} & Generated & 3.353 & 3.338 & 0.082 & \textbf{0.921} & 16.981 & 2.416 & 3.189 & 0.012 & \textbf{0.950} & 14.249 \\
      & HALLO \cite{hallo} & Generated & 6.542 & 2.052 & \underline{0.182} & 0.525 & 19.574 & 5.298 & 2.301 & 0.104 & 0.320 & 15.887 \\
      \multicolumn{1}{l}{\multirow{2}{*}{}} &
      \multicolumn{1}{l}{\cellcolor[gray]{0.95}Ours-image} & \cellcolor[gray]{0.95}Generated & \cellcolor[gray]{0.95}7.095 & \cellcolor[gray]{0.95}2.628 & \cellcolor[gray]{0.95}\textbf{0.318} & \cellcolor[gray]{0.95}\underline{0.912} & \cellcolor[gray]{0.95}\underline{24.722} & \cellcolor[gray]{0.95}\textbf{6.419} & \cellcolor[gray]{0.95}\underline{2.139} & \cellcolor[gray]{0.95}\textbf{0.249} & \cellcolor[gray]{0.95}0.812 & \cellcolor[gray]{0.95}\underline{22.044} \\
      \multicolumn{1}{l}{} &
      \multicolumn{1}{l}{\cellcolor[gray]{0.95}Ours-video} & \cellcolor[gray]{0.95}From ref & \cellcolor[gray]{0.95}\textbf{7.289} & \cellcolor[gray]{0.95}\underline{1.991} & \cellcolor[gray]{0.95}\underline{0.212} & \cellcolor[gray]{0.95}0.909 & \cellcolor[gray]{0.95}\textbf{30.533} & \cellcolor[gray]{0.95}\underline{6.351} & \cellcolor[gray]{0.95}\textbf{1.366} & \cellcolor[gray]{0.95}\textbf{0.206} & \cellcolor[gray]{0.95}\underline{0.869} & \cellcolor[gray]{0.95}\textbf{29.870} \\
      \bottomrule
    \end{tabular}%
    }
  \label{tab:comparison}
\end{table*}
\section{Experiments}
\begin{table}[t!]
  \centering
  \caption{Cross-lingual evaluation on \texttt{Voxceleb2} \cite{vox2} dataset.}
  \resizebox{0.75\columnwidth}{!}{
    \begin{tabular}{lcccc}
      \toprule
       Methods &  LSE-C$\uparrow$ & LMD$\downarrow$ & Blink$\uparrow$ & CPBD$\uparrow$ \\
      \midrule
       GT & 5.554 & 0.000 & 0.180 & 0.688 \\
      \midrule
      AVCT \cite{avct} & 4.173 & 3.404 & 0.090 & 0.373 \\
      Audio2Head \cite{audio2head} & 5.248 & 3.113 & 0.029 & 0.307 \\
      MakeItTalk \cite{makeittalk} & 4.114 & 1.743 & 0.108 & 0.608 \\
      SadTalker \cite{sadtalker} & 5.527 & 2.700 & 0.116 & 0.330 \\
      Styletalk \cite{styletalk} & 1.835 & 3.399 & 0.187 & 0.432 \\
      IPLAP \cite{iplap} & 4.362 & 3.217 & 0.007 & 0.344 \\
      PDFGC \cite{pdfgc} & 5.659 & 2.395 & 0.032 & 0.670 \\
      \midrule
      Dreamtalk \cite{dreamtalk} & 4.745 & 2.669 & 0.007 & 0.215 \\
      Aniportrait \cite{aniportrait} & 2.455 & 3.188 & 0.025 & \textbf{0.895} \\
      HALLO \cite{hallo} & \underline{4.933} & \textbf{2.340} & \underline{0.139} & 0.189 \\
       \cellcolor[gray]{0.95}Ours & \cellcolor[gray]{0.95}\textbf{6.299} & \cellcolor[gray]{0.95}\underline{2.591} & \cellcolor[gray]{0.95}\textbf{0.165} & \cellcolor[gray]{0.95}\underline{0.672} \\
      \bottomrule
    \end{tabular}%
  }
  \label{tab:comparison}
\end{table}

\subsection{Implementation Details and Metrics}
\noindent\textbf{Datasets.} Existing audio-visual datasets \cite{hdtf,vox2} predominantly focus on English content, limiting the generalization capability to other languages, particularly those with distinct phonetic characteristics. Mandarin Chinese, as one of the most widely used languages on the Internet, contains unique phonemes absent in English, such as retroflex consonants (e.g., "zh, ch, sh") and tonal variations (e.g., "ma" with four different tones representing "mother", "hemp", "horse", and "scold"). To address this limitation, we introduce \textbf{\texttt{CHDTF} (Chinese High-Definition Talking Face Dataset)}, comprising 1,537 high-quality videos (39.1 hours) and most videos anchored at 1080p resolution. The dataset's phonetic diversity and professional recording quality make it valuable for both Chinese-specific applications and cross-lingual generalization. For comprehensive evaluation, we utilize \texttt{HDTF} \cite{hdtf} (300+ videos, 15.8 hours) for English content and \texttt{Voxceleb2} \cite{vox2} for in-the-wild cross-lingual testing. All videos are preprocessed to 256×256 resolution and audio standardized to 16kHz mono channel format.

\noindent\textbf{Implementation Details.} We implement our framework in PyTorch and train it using the Adam optimizer with a learning rate of 1e-4 and batch size of 32. The region-specific dimension selection is determined by the ratio of the maximum temporal variation to the standard deviation, resulting in dimensions $K_{lip} = 13$ and $K_{eye} = 8$. The diffusion process uses T = 400 timesteps, and the local convolution window size $k$ is set to 5. The sync loss weight $\lambda$ is set to 0.01 to balance different optimization objectives. \footnote{Detailed experimental settings and datasets are provided in the supplementary material.}

\noindent\textbf{Metrics.} We adopt peak signal-to-noise ratio (PSNR) ~\cite{PSNR} and cumulative probability of blur detection (CPBD) \cite{cpbd} to assess the image quality and clarity. To evaluate lip synchronization, we use the lip-sync error confidence score (LSE-C) from \cite{syncnet}, and the landmark distance of the mouth (LMD) \cite{atvg}. To assess overall facial naturalness, we measure the blinking frequency. Following human physiological standards where normal blinking rates range from 0.28 to 0.45 blinks/s \cite{blink}, we use blink rate (Blink) as a naturalness metric, which provides insight into the realism of generated facial movements.

\begin{figure*}[!ht]
  \centering
  \includegraphics[width=0.95\textwidth]{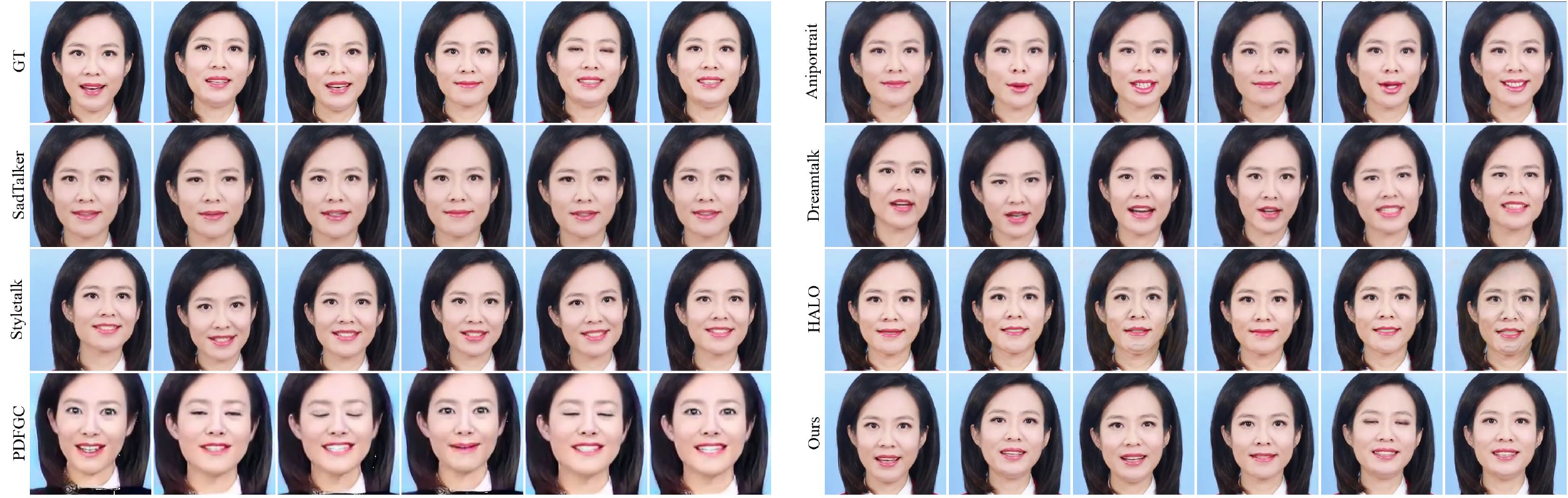}
  \caption{Qualitative comparison with state-of-the-art methods. Our approach demonstrates superior performance in lip synchronization, natural facial expressions, and temporal consistency.}
  \label{fig:qualitative}
  \vspace{-3mm}
\end{figure*}

\subsection{Experimental Results}
\noindent\textbf{Baselines.} We compare our method with ten state-of-the-art approaches, categorized into two groups: (1) \textit{Deterministic methods}: MakeItTalk \cite{makeittalk}, Audio2Head \cite{audio2head}, AVCT \cite{avct}, IPLAP \cite{iplap}, SadTalker \cite{sadtalker}, Styletalk \cite{styletalk}, and PDFGC \cite{pdfgc}. (2) \textit{Diffusion-based methods}: Dreamtalk \cite{dreamtalk}, Aniportrait \cite{aniportrait}, and HALLO \cite{hallo}. The reported results were obtained using the publicly available code of these methods. (note: the input speaker of ours-video, IPLAP, and PDFGC are videos, while the other methods are a single image.)

\noindent\textbf{Quantitative Results.} We conduct comprehensive evaluations on \texttt{HDTF} \cite{hdtf} (English), \texttt{CHDTF} (Chinese), and \texttt{Voxceleb2} \cite{vox2} (multi-lingual) datasets to assess the effectiveness of our method in cross-lingual talking face generation. Our approach demonstrates superior performance across multiple aspects: achieving state-of-the-art LSE-C scores (6.299 on \texttt{Voxceleb2} vs. baseline average~4.5) in cross-lingual scenarios, maintaining physiologically plausible blink rates (0.165 on \texttt{Voxceleb2}) through semantic disentanglement, and exhibiting consistent lip synchronization across languages (LSE-C: 7.289/6.419/6.299 on \texttt{HDTF}/\texttt{CHDTF}/\texttt{Voxceleb2}). Notably, while Aniportrait achieves high CPBD (0.895) through its Stable Diffusion backbone, our method maintains competitive image clarity (CPBD: 0.672) while significantly outperforming in temporal coherence and lip synchronization, demonstrating a better balance between spatial fidelity and motion dynamics. These comprehensive results validate our approach's effectiveness in addressing the core challenges of cross-lingual talking face generation through semantic disentanglement and spatio-temporal aware modeling.

\noindent \textit{What is the efficiency of our method compared to others?}

For a 53-second test video, our method (31.56 FPS) is significantly faster than AniPortrait (0.92 FPS) and HALLO (0.28 FPS), enabling real-time generation.


\begin{figure}[!t]
  \centering
  \includegraphics[width=0.95\columnwidth]{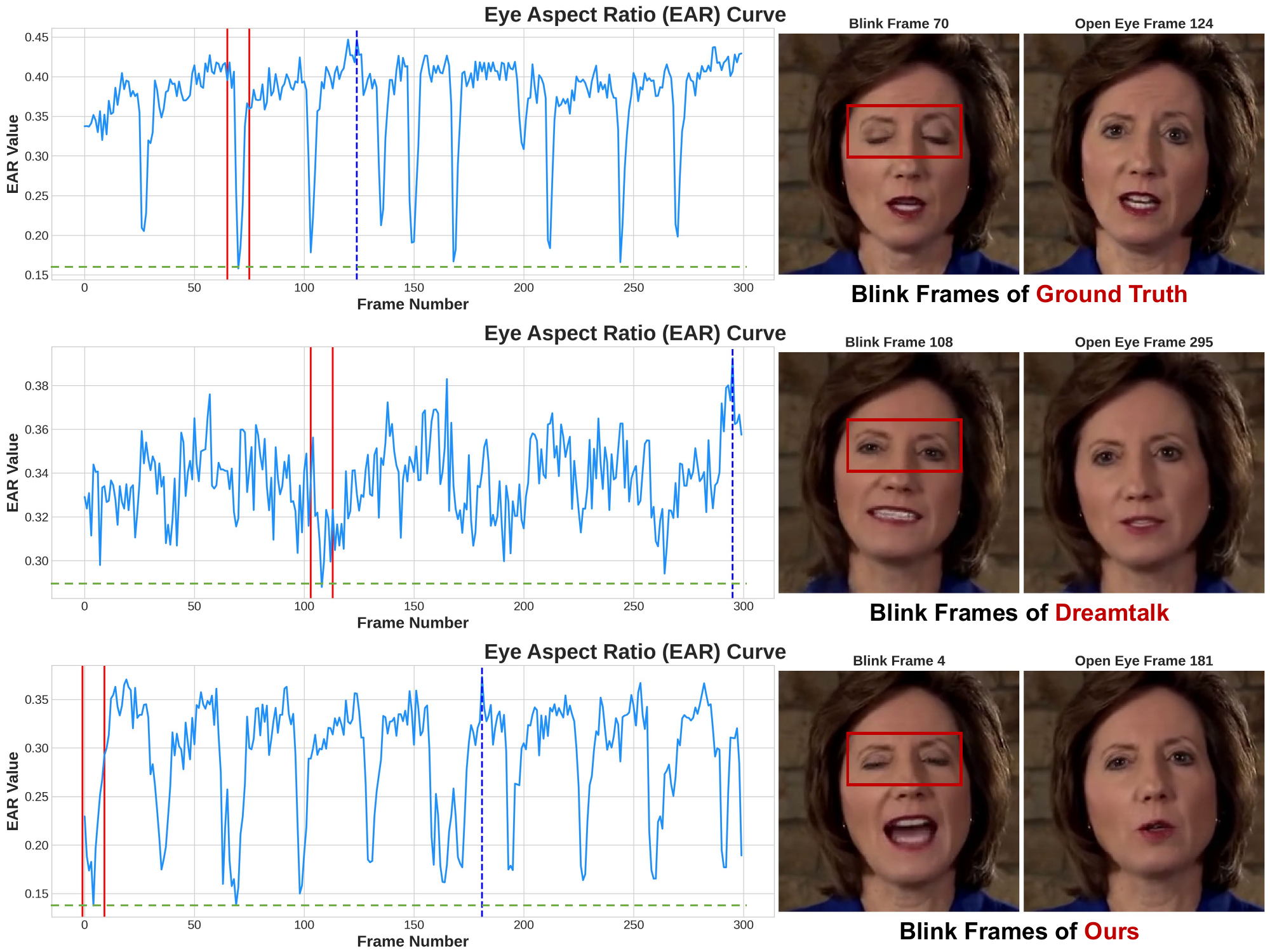}
  \caption{Eye aspect ratio analysis over time. Our method generates physiologically plausible blinking patterns compared to baselines.}
  \label{fig:eye_pattern}
\end{figure}

\begin{figure}[!t]
  \centering
  \includegraphics[width=\columnwidth]{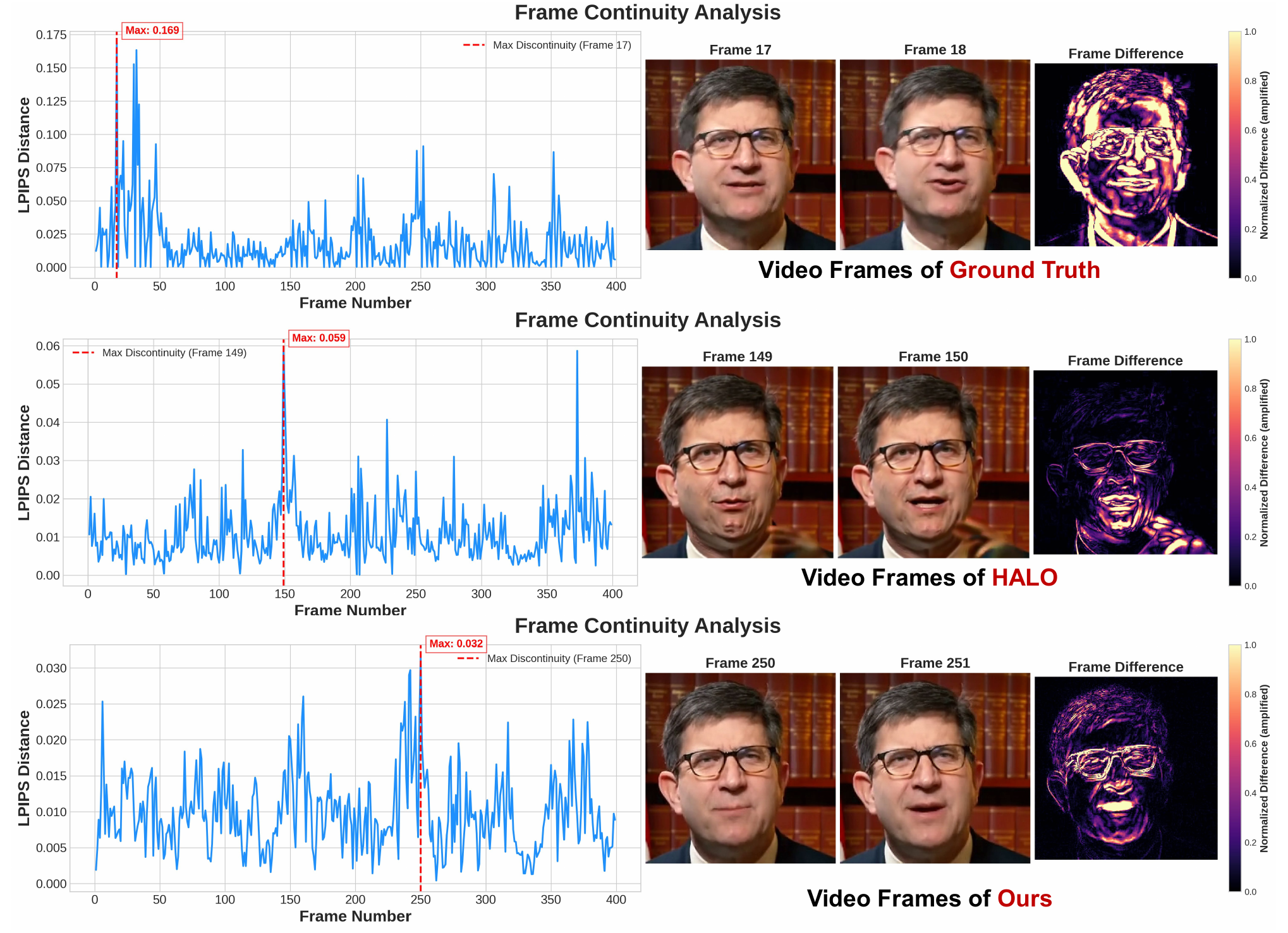}
  \caption{LPIPS \cite{lpips} differences between consecutive frames.}
  \label{fig:temporal}
\end{figure}

\begin{table}[t]
\centering
\caption{Comparison of mean opinion scores (MOS) with standard deviations across different methods. }
\resizebox{\columnwidth}{!}{
\label{tab:mos_comparison}
\begin{tabular}{l|cccc|>{\columncolor{gray!20}}c}
\toprule
\textbf{MOS} & SadTalker \cite{sadtalker} & DreamTalk \cite{dreamtalk} & AniPortrait \cite{aniportrait} & HALLO \cite{hallo} & \textbf{Ours} \\
\midrule
Lip Synchronization    & 3.44 & 3.47 & 2.52 & \underline{3.60} & \textbf{3.84} \\
Expression Naturalness & 3.37 & 3.32 & 2.75 & \underline{3.55} & \textbf{3.93} \\
Temporal Coherence    & \underline{3.73} & 3.66 & 3.43 & 3.20 & \textbf{3.85} \\
\bottomrule 
\end{tabular}
}
\label{tab:mos_comparison}
\end{table}
\noindent\textbf{Qualitative Results.} 
We conduct qualitative comparisons on randomly sampled test videos from the \texttt{CHDTF} dataset (Fig. \ref{fig:qualitative}). While HALLO exhibits unnatural wrinkle artifacts and other baselines struggle with lip synchronization, our method demonstrates superior performance in both lip-sync accuracy and expression naturalness. Analysis of the aspect ratio of the eye (Fig. \ref{fig:eye_pattern}) validates the effectiveness of our semantic disentanglement framework, showing physiologically plausible blinking patterns compared to Dreamtalk's unnatural eye movements. Frame-wise temporal consistency evaluated by LPIPS \cite{lpips} differences between consecutive frames (Fig. \ref{fig:temporal}) shows that HALLO exhibits significant frame-to-frame feature variations even during subtle head movements, while our approach maintains consistent visual features across adjacent frames. These observations are supported by our user study (Table \ref{tab:mos_comparison}), where our method achieves the highest scores in lip synchronization (3.84), expression naturalness (3.93), and temporal coherence (3.85). 


\subsection{Ablation Study}

\begin{table}[t!]
  \centering
  \caption{Ablation study of different components in our method. }
  \resizebox{0.9\columnwidth}{!}{
    \begin{tabular}{lcccc}
      \toprule
      Methods & LSE-C$\uparrow$ & LMD$\downarrow$ & Blink$\uparrow$ & CPBD$\uparrow$ \\
      \midrule
      Ours w/o \textsc{temporal} & 5.730 & 2.216 & 0.055 & 0.811 \\
      Ours w/o \textsc{disentangle} & \underline{5.993} & 2.243 & 0.000 & \textbf{0.819} \\
      Ours w/o \textsc{hierarchical} & 5.955 & \underline{2.193} & 0.006 & 0.807 \\
      Ours w/o \textsc{local} & 5.846 & 2.276 & 0.053 & 0.808 \\
      Ours w/o \textsc{CHDTF} & 5.389 & 2.326 & 0.154 & 0.810 \\
      Ours w/o \textsc{syncloss} & 4.470 & 2.380 & \underline{0.195} & 0.812 \\
      \cellcolor[gray]{0.95}Ours full & \cellcolor[gray]{0.95}\textbf{6.419} & \cellcolor[gray]{0.95}\textbf{2.139} & \cellcolor[gray]{0.95}\textbf{0.249} & \cellcolor[gray]{0.95}\underline{0.812} \\
      \bottomrule
    \end{tabular}%
  }
  \label{tab:ablation}
\end{table}

 We conduct ablation experiments to evaluate key components: multi-scale temporal convolution (\textsc{temporal}), 3DMM parameter disentanglement (\textsc{disentangle}), hierarchical region-aware spatial attention (\textsc{hierarchical}), local spatio-temporal correlations (\textsc{local}), \textsc{CHDTF} dataset training, and lip sync loss (\textsc{syncloss}). Table \ref{tab:ablation} shows that: 1) Removing \textsc{disentangle} and \textsc{hierarchical} leads to complete failure in eye blinking (Blink: 0.000, 0.006), underscoring their essential role in achieving fine-grained control over distinct facial regions, a key limitation of traditional 3DMM-based methods; 2) \textsc{temporal} and \textsc{local} modules improve lip synchronization (LSE-C from 5.730, 5.846 to 6.419); 3) \textsc{syncloss} and \textsc{CHDTF} enhance lip sync performance (LSE-C drops to 4.470, 5.389 when removed). Note that CPBD values remain similar (0.807-0.819) due to consistent image rendering.

\section{Conclusion}

In this paper, we presented a novel framework for cross-lingual talking face generation that effectively addresses two fundamental challenges: semantic entanglement in facial representations and isolated processing of spatial-temporal features. Through our semantic disentanglement approach and hierarchical diffusion architecture, we achieve precise control over distinct facial regions while maintaining temporal coherence. The CHDTF dataset enriches the training data and enables cross-lingual evaluation. Extensive experiments show superior performance over existing methods in lip synchronization, expression naturalness, and temporal stability.

\bibliographystyle{IEEEbib}
\bibliography{icme2025references}

\end{document}